\documentclass[conf]{new-aiaa}
\usepackage[utf8]{inputenc}
\usepackage{subcaption}
\usepackage[]{graphicx}  % remove 'demo' option for your real document
\usepackage{amsmath}
\usepackage[version=4]{mhchem}
\usepackage{siunitx}
\usepackage{longtable,tabularx}
\usepackage[linesnumbered,ruled]{algorithm2e}
\usepackage{booktabs}
\usepackage{floatrow}
\usepackage{soul}
\graphicspath{{imgs/}}

\title{Automatic Classification of Roof Shapes \\ for Multicopter Emergency  Landing Site Selection \\ (Extended Abstract)}

\author{Jeremy~D.~Castagno\footnote{J. D. Castagno is with the Department of Aerospace Engineering, University of Michigan, Ann Arbor, MI, 48109 USA e-mail: jdcasta@umich.edu.} and Ella~M.~Atkins\footnote{E. M. Atkins   is with the Department
of Aerospace Engineering, University of Michigan, Ann Arbor,
MI, 48109 USA e-mail: ematkins@umich.edu.}}
\affil{University of Michigan, Ann Arbor, MI, 48109}
\begin{document}

\maketitle

% No separate abstract is needed in an extended abstract
%\begin{abstract}
%Geographic information systems (GIS) now provide accurate maps of terrain, roads, waterways, and building footprints and heights. Aircraft, particularly small unmanned aircraft systems, can exploit additional information such as building roof structure to improve navigation accuracy and safety particularly in urban regions.  This paper proposes a method to automatically label building roof shape types. Satellite imagery and LIDAR data from Witten, Germany are fed to convolutional neural networks (CNN) to extract salient feature vectors. Supervised training sets are automatically generated from pre-labeled buildings contained in the OpenStreetMap database. Multiple CNN architectures are trained and tested, with the best performing networks providing a condensed feature set for support vector machine and decision tree classifiers. Satellite and LIDAR data fusion is shown to provide greater classification accuracy than through use of either data type individually.
%\end{abstract}
\vspace{-0.5cm}
\section{Introduction}
\lettrine{G}{eographic} information system (GIS) data is openly available for a variety of applications. Data on terrain height has historically been available; high-accuracy labelled data is now also available, e.g., indicating building footprints and heights.  Unmanned Aircraft Systems (UAS) offer low-cost airborne data collection and small payload transport capabilities to support a variety of missions including but not limited to urban mapping \citep{thomas2003comparison}. 
%Archived data can support realistic three-dimensional (3D) simulation environments \citep{stoor2006urban} as well as accurate maps to support map-based localization.  
Routine small UAS operation over urban regions requires mitigation of risk posed to overflown people and property.  Small UAS are envisioned to operate at low altitudes, below the airspace occupied by manned aircraft.  In a congested environment, the low-flying UAS will operate near buildings thus will have little time or space to execute a "safe ditching" (emergency landing) given an unrecoverable anomaly or failure \citep{ochoa2017fail}.   Concise and accurate information of nearby landing sites, including building rooftops, is needed in these situations. Databases such as OpenStreetMap (OSM) (\url {https://www.openstreetmap.org/}) and Mapbox (\url {https://www.mapbox.com/}) provide open data on infrastructure as well as terrain.  Roof geometry is offered as a possible field in OSM that could provide information for a small UAS; however, roof geometry and architecture (type) information is often missing or incomplete in these existing databases. This paper provides missing roof type classification information from publicly-available GIS data.

This paper fuses satellite imagery and LIDAR data through multiple stages of machine learning classifiers to accurately characterize building rooftops.  With these results, roof geometries worldwide can be stored in an easily-accessible format for UAV and other applications. Supervised training datasets are automatically generated by combining OSM, satellite, and LIDAR data. The resulting annotated dataset provides individual satellite image and LIDAR image representations for each building roof. Roof shapes are automatically labelled through a novel combination of convolutional neural networks (CNNs) using these roof images from satellite and LIDAR data sources.  Transfer learning is employed in which pre-trained CNN model architectures and hyper-parameters are fine-tuned and tested. The best performing CNN for both satellite and LIDAR data inputs is used to extract a reduced feature set which is then fed into support vector machine (SVM) and random forest classifiers to extract a single roof geometry decision.  Validation and test set accuracies are evaluated over a suite of different classifier options. Imagery and roof classification data from Witten, Germany is used in this work based on prior availability of human-input (crowd-sourced) roof geometry classification data for buildings throughout the region.

% Automatically generating supervised training sets using crowd sourced data through OSM
% Performing a much larger analysis of MULTIPLE of CNN architectures which are pretrained
% Fusing the LIDAR and RGB features into non-linear decision models to achieve greater accuracy results
% Performing feature extraction that is later fed to MULTIPLE SVM and random forest classifiers

% The paper is structured as follows.  First, a review of GIS data sources and prior roof geometry classification work is presented.  Next, background in machine learning and data extraction methods is provided.  Specific methods to extract data for input to this paper's machine learning feature extraction and classification system are presented, followed by a description of training, validation, and test runs performed.  Statistical accuracy results are presented followed by a discussion and conclusions.

\section{Background}
% This section summarizes related work on GIS data sources, convolutional neural networks (CNNs), and their application to feature extraction.
% %This section provides background on related work and fundamental classification approaches applied in this paper. First, GIS data sources and related work in extracting roof geometries is summarized.  Next, convolutional neural networks (CNNs) and their application to feature extraction is reviewed.
% \vspace{-0.1cm}
% \subsection{Roof Geometry Classification}

% % Too general -- focus on specifics. -Ella
% % Many methods exist for the automatic determination of roof geometry, depending greatly on the available data, the level of detail desired, as well as the many algorithms of choice.  
Satellite color images and 3D point cloud data from airborne Light Detection and Ranging (LIDAR) sensors provide complementary roof information sources. High resolution satellite images offer rich information content and are generally available worldwide.  However, extracting 3D building information from 2D images is difficult due to occlusion, poor contrast, shadows, and skewed image perspectives \citep{zhou2014seamless}. Raw LIDAR sensors provide depth and intensity measurements generating a 3D point cloud capturing the features of roof shapes, yet LIDAR does not offer other world feature information from ambient lighting intensity and color. LIDAR point cloud data is often processed and converted to digital surface models (DSM) representing the top surface layer of any terrain. 

The UAV localization and emergency landing application targeted by this paper only requires a simple classification of a building roof shape.   In fact, complex model representations are undesirable given that a UAV emergency landing plan to a flat rooftop, for example, would be computed by a low-power lightweight embedded processor. Classical machine learning algorithms such as support vector machines (SVM), logistic regression, and decision trees are often used in these classification scenarios but invariably face computational complexity challenges caused by the high dimensionality found in these GIS data sources. Deep learning, through the use of techniques such as convolutional neural networks (CNN), have demonstrated the ability to accurately and robustly classify high dimensional data sources such as camera images \citep{schmidhuber2015deep}. The GIS community has begun to apply CNNs to roof identification.  Perhaps most closely related to this paper, Alidoost and Arefi (2016) trained CNNs using satellite (RGB) and digital surface map (DSM) images to label basic roof shapes \citep{alidoost2016knowledge}.  However the final predicted roof shape was simply taken as the highest probability result between the two models. 
%Two successful strategies were used in order to combine both RGB and DSM inputs for a final label prediction. The first strategy created two CNN models by fine-tuning a pre-trained CNN for both RGB and DSM inputs. The final predicted roof label was simply taken as the highest probability between the two models. The second (and best) strategy remained the same, except the training of the DSM CNN was initialized to the fine-tuned RGB model, leading to greater model accuracy and reduced training time. 
% %The third strategy employed Principal Component Analysis on the combined RGB-D dataset, eventually whitening the data as input for training another CNN.
Patrovi et. al \citep{partovi2017roof} fine-tuned a CNN using patched satellite images of building roof-tops.  Using the fine-tuned CNN, the authors extracted high-level features of images as inputs to a second-stage SVM classifier. This paper adopts an analogous two-stage processing approach to roof classification with the addition of LIDAR and satellite feature fusion. Specifically, this fusion allows the creation of a nonlinear decision function that exploits the strengths of each modality.  Note that the full paper will contain additional background on CNNs in the context of RGB and LIDAR feature detection and classification problems.

\section{GIS Roof Data Extraction and Processing}

To classify roof geometries, a large annotated data set must be processed and randomly split into distinct training, validation, and testing subsets. This paper relies on a fusion of OSM, satellite imagery, and airborne LIDAR data sources to generate this data set. Each building roof is classified based on corresponding satellite (RGB) and LIDAR images of the roof top.  

% {\color{blue} Edits:  use capitalized LIDAR everywhere.  Use consistent tense -- ideally present when possible and past when describing what was done as a precursor to this paper's methods.  Performing, not preforming.}

%The training image subset is fed into a number of distinct CNN architectures in randomized order.  The validation set is used to determine the best-performing CNN network. The best CNN is then used to extract high level features of the same training set images to be used as training feature inputs to SVM and random forest classifiers.  CNN, SVM, and random forest classifiers are then assessed based on their performance over the test data set. Details of pre-classified image set generation, CNN training, and feature extraction for SVM and decision tree second-stage classifiers are presented in the following sections.

\subsection{Classified Image Set Generation}

Generation of an annotated roof data set requires three data sources for each building: satellite RGB imagery, airborne LIDAR data, and building outlines with corresponding roof labels (from manual classification).  All three of these data sources must be properly geo-referenced so they can be fused together. Care must be taken to select a geographic area where data sources for all of these items are present, as well as a sufficiently dense number of pre-labeled buildings. OpenStreetMap (OSM) provides the necessary building outlines in almost all geographic regions, however the associated roof shape label is often not complete. Some geographic regions (e.g. Germany) are more likely to have a denser collection of labeled roof shapes through a higher volunteer involvement.  Once the appropriate data sources have been found, the methods described below can be employed to extract satellite and LIDAR images for each building.

%  {\color{blue} More sparsely labeled than what other data source?}

Satellite, LIDAR, and OSM data sources have their own spatial reference systems (SRS). The SRS defines a map projection and determines the transformations needed to convert to a different SRS.  These reference systems are uniquely identified though a spatial references system identifier (SRID) which designates an authority and an identifier.  For example, the  European Petroleum Survey Group (EPSG) can be used to specify SRID's. OSM chooses to store building outlines as polygons, with each vertex stored in WGS84 (EPSG:4326). Satellite images from common map vendors (ArcGIS, Bing, Google) often use WGS84 / Pseudo-Mercator (EPSG:3857). LIDAR data is usually stored in a region-specific SRS; data for Witten, Germany uses EPSG:5555. To convert a point stored in one SRS to another, a program specialized in these transformations, such as \texttt{proj.4}, must be used.  OSM building polygons are transformed to their LIDAR and satellite counterpart coordinate systems so that the building outlines are consistent.

% A common spatial reference must be chosen in order to integrate these three data sources. This paper uses the spatial reference system (SRS) of the LIDAR data set because its SRS was specifically chosen to best represent the geographic area of interest. OSM chooses to store their data in WGS84 (EPGS:4326) coordinates, while the satellite images from common vendors (ArcGIS, Bing, Google Maps) often store data in the  projected 2D coordinate system of EPGS:3857. After projecting to a common coordinate system, these data sources can be presumed correlated for processing. All images generated must be resized to the square dimensions specified by the CNN architecture to be trained, validated, and tested.

\subsubsection{LIDAR Image Construction}
A bounding box (BBOX) is constructed from the polygon building outline provided by OSM.  This BBOX is used first to quickly filter out points in the LIDAR data set not related to the building of interest.  The resulting subset of points is filtered again using the polygon roof outline, resulting in only points encapsulated in the building outline. At this time, the 3D LIDAR point cloud is noisy and may contains undesirable points, especially near the edges of the outline (i.e. wall hits).  To remove these outliers, the points' $z$-coordinates are used to analyze their distribution. A robust metric for outlier removal is to use the median absolute deviation (MAD) and construct a modified \texttt{z-score} that measures how deviant each point is from the MAD  \citep{iglewicz1993detect}. This method only applies to unimodal distributions; however not all buildings height are distributed as such. For example, there exist complex flat buildings that contain multiple height levels resulting in a multimodal distribution. To distinguish these buildings the dip test statistic is employed which measures multi-modality in a sample distribution \citep{hartigan1985algorithm}. Any building with a dip statistic less than .04 or with a p-value greater than .2 is considered unimodal, and outlier removal is performed.
% as shown in Algorithm \ref{alg:filter}.

% \begin{algorithm}
%     \SetKwInOut{Input}{Input}
%     \SetKwInOut{Output}{Output}

%     \Input{Collection of 3D points, $A$}
%     \Output{Filtered 3D point cloud $B$}
%     $Z$ = $A_z$ \\
%     $B$ = $\emptyset$ \\
%     dip, p-value = $\operatorname{diptest}(Z)$ \\
%     \eIf{$\text{dip} \leq .04 \lor \text{p-value} \geq .2$}{
%         ${\operatorname {MAD} =\operatorname {median} \left(\ \left|Z_{i}-\operatorname {median} (Z)\right|\ \right)}$ \\
%         \For{$p$ in $A$}{
%             \texttt{diff} = $|p_z - \operatorname {median} (Z) |$ \\
%             \texttt{z-score} = $0.6745 \cdot \texttt{diff} / \operatorname {MAD}$ \\
%             \If{$\texttt{z-score} \leq 3.0$}{
%                 $B$ = $B$ + $p$
%              }
%         }
%     }{
%         $B$ = $A$
%     }
%     return $B$
%     \caption{Filtering of 3D LIDAR point cloud using Medium Absolute Deviation}
%     \label{alg:filter}
% \end{algorithm}

\begin{figure}[t]

\centering
   \begin{subfigure}[b]{0.45\textwidth}
   \includegraphics[width=.95\linewidth]{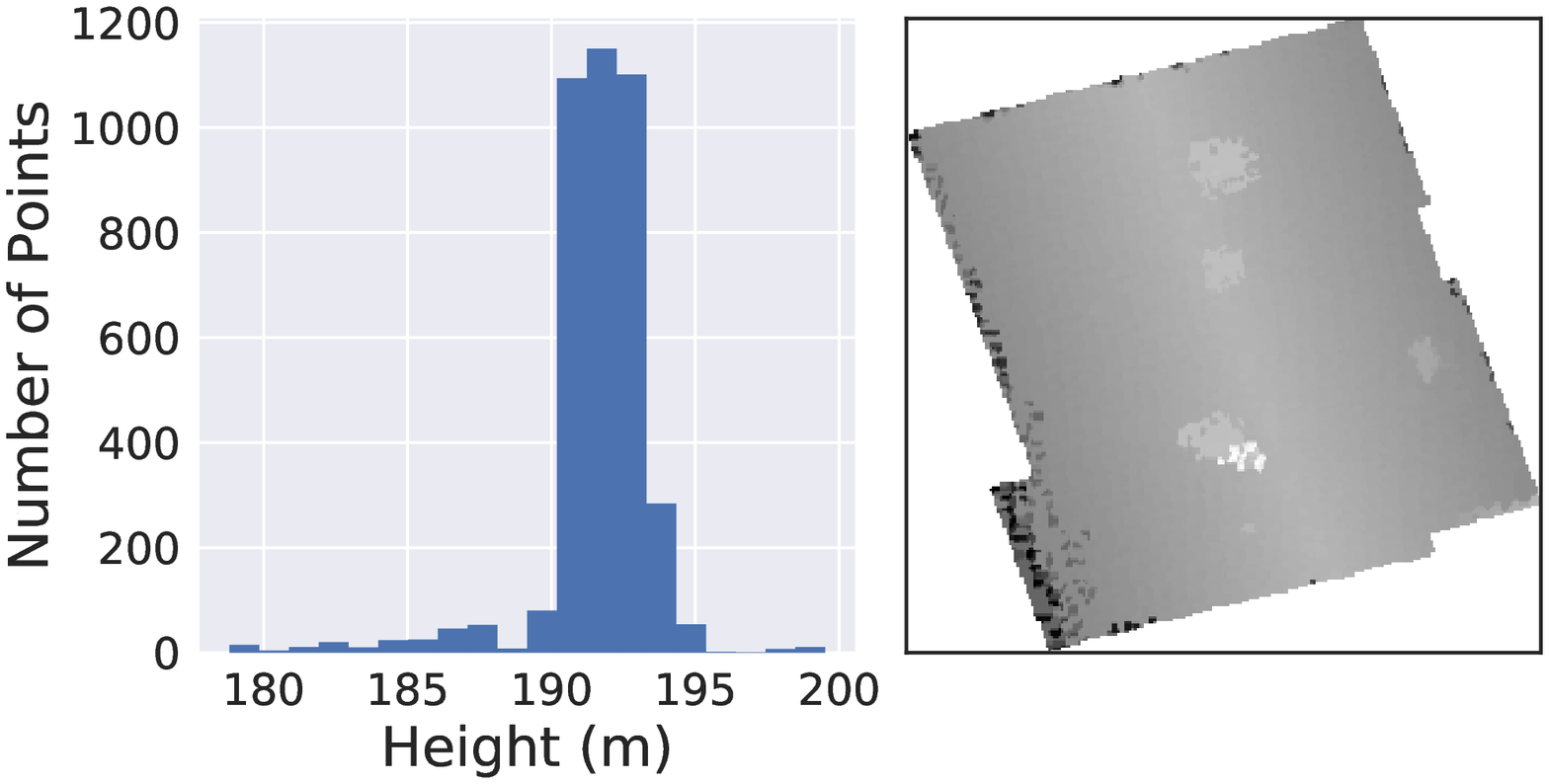}
   \caption{}
   \label{fig:lidar_filt_before} 
\end{subfigure}
\begin{subfigure}[b]{.45\columnwidth}
   \includegraphics[width=.95\linewidth]{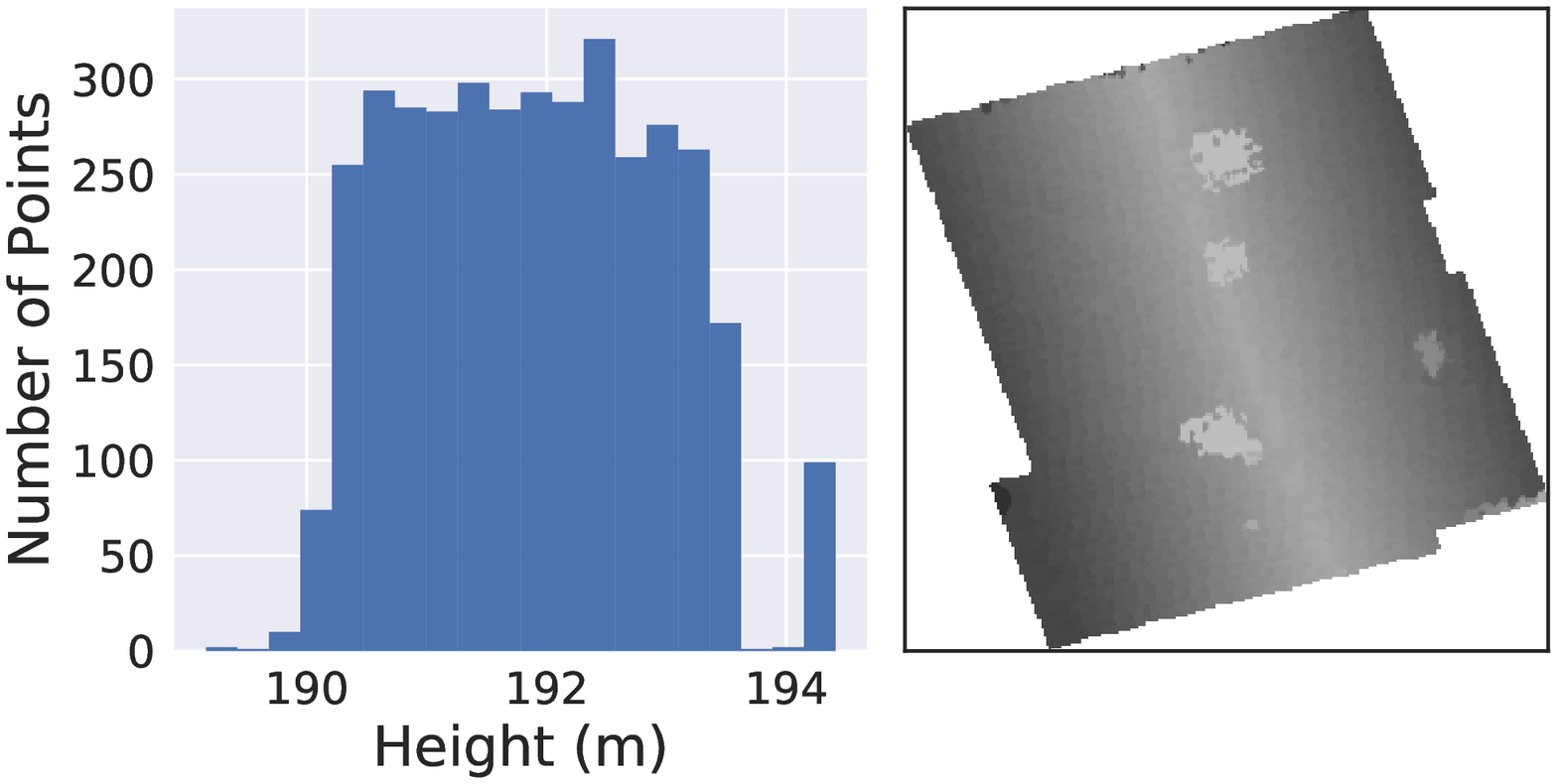}
   \caption{}
   \label{fig:lidar_filt_after}
\end{subfigure}

\caption[Results of LIDAR Filtering]{LIDAR data of a gabled roof.  Histogram of height distribution and generated image (a) before filtering and (b) after filtering, using median absolute deviation.}
\label{fig:lidar_filt}
\end{figure}

Once LIDAR point extraction is complete, the points are projected onto a plane, creating a 2D grid that takes the value of each point's height information. The 2D grid world dimensions are the same as the bounding box of the buildings, with the discrete grid size being the desired square image resolution. Grid points use interpolation of nearest neighbor if no point is available. Afterward this grid is converted into a grayscale image, where each value is scaled from 0-255 with higher values whiter and lower areas darker.  Examples of this filtering are shown in Figure \ref{fig:lidar_filt}. The CNNs used in this paper require the grayscale LIDAR data be converted to a three-channel RGB image by duplicating the single channel across all three color channels. This final image is referred to as the LIDAR image.

\subsubsection{Satellite Image Construction}

Satellite RGB (red-green-blue) images must have a a minimum ground resolution of 1m, however sub-meter resolution is recommended. It is preferable that the imagery be ortho-rectified to remove image tilt and relief effects. Ideally the OSM building polygon can be used to completely \emph{stamp} out a roof shape image. However, if the aforementioned issues are present in the image, it is unlikely that the OSM polygon will exactly match the building outline in the image. This was the case when using ArcGIS world imagery for the city of Witten, Germany. To work around these issues, an enlarged crop can be made around the building.  The enlarged crop is produced by orthogonally expanding each edge of the building polygon by a configurable constant, and then using the bounding box of the new polygon as the identifying stamp.  After experimentation, this configurable constant was set to three meters when processing the Witten data set. After the image is produced, the image is resized to the square image resolution required by the CNN.  
% Figure \ref{fig:rgb_overlay} shows the original OSM building outline (red shade) overlaid on a satellite image, and the expanded polygon bounding box in cyan. The resulting generated image is shown in Figure \ref{fig:rgb_building}. This final images is referred to as the RGB image.

% \begin{figure}[t]
%     \centering
%     \begin{subfigure}[t]{0.5\columnwidth}
%         \centering
%         \includegraphics[height=1.2in]{rgb_building_overlay.png}
%         \caption{}
%         \label{fig:rgb_overlay}
%     \end{subfigure}%
%     \hfill
%     \begin{subfigure}[t]{0.5\columnwidth}
%         \centering
%         \includegraphics[height=1.2in]{rgb_building.png}
%         \caption{}
%         \label{fig:rgb_building}
%     \end{subfigure}
%     \caption{(a) OSM building outline, red shade, overlaid on satellite image. Enlarged crop area shown in cyan shading. (b) Final image produced.}
% \end{figure}

\subsection{CNN Architectures and Training}

\begin{figure*}[ht!]
\centering
\includegraphics[width=0.8\textwidth]{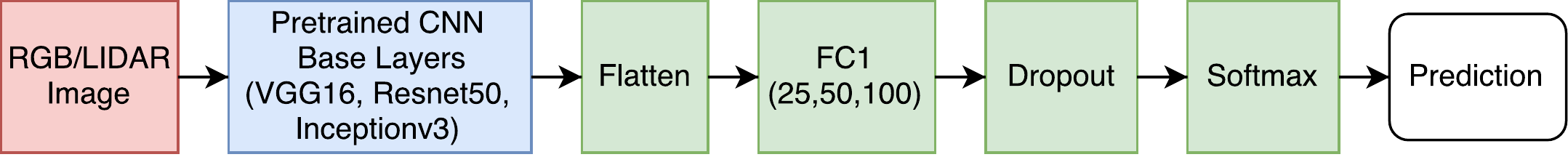}
\caption{CNN Architecture Templates}
\label{fig:cnn_architectures}
\end{figure*}

The CNN base architectures chosen for experimentation are \texttt{VGG16} \citep{DBLP:journals/corr/SimonyanZ14a}, \texttt{Resnet50} \citep{He2015}, and \texttt{Inceptionv3} \citep{DBLP:journals/corr/SzegedyVISW15}.  All three of these architecture structures are distinct; when trained and tested on ImageNet they received "top 5" accuracy scores of 89.8\%, 92.8\%, and 93.9\% respectively. Each CNN makes use of successive convolutional blocks to generate a final feature map (referred to as the base layers) which are subsequently used by downstream fully-connected layers to make a 1000 categorical prediction (referred to as the top layers). The top layers are domain specific and are not needed for the roof classification task. Three new architecture templates are constructed using the pre-trained CNN base layers: a flattening layer, a fully connected layer (FC1), and a softmax prediction layer as shown in Figure \ref{fig:cnn_architectures}. Additionally the size of FC1 is also varied over vector lengths of 25, 50, and 100.  These models are then subsequently trained individually on the RGB and LIDAR images.

% {\color{blue} Use courier text for algorithm and variable names.  Example: \texttt{Inceptionv3}.  Use common capitalization also on each reference to each algorithm.}

% Have to put this figure here.....

Training will initialize the weights for the base layers with their respective parent architecture. The optimizer chosen for gradient descent is Adadelta \citep{DBLP:journals/corr/abs-1212-5701} for its ability to effectively adjust the learning rate automatically for individual weights; this optimizer is kept consistent for all architectures and training sessions. The option of freezing initial layers is exploited with a variable number of frozen layers chosen. When layer 11 is said to be frozen, this means all previous layers, (1-11), are frozen during training. All base architectures and tested hyper-parameters are shown in Table \ref{table:cnn_arch}.  

% \usepackage{booktabs}

\begin{table}
\begin{floatrow}
\ttabbox
{\caption{CNN Architectures and Hyperparameters}\label{table:cnn_arch}}{
\begin{tabular}{@{}lll@{}}
\toprule
Base CNN Model & FC1 Size  & Frozen Layers \\ \midrule
VGG16          & 25,50,100 & 11, 15        \\
Resnet50       & 25,50,100 & 50, 80        \\
InceptionV3    & 25,50,100 & 18, 87        \\ \bottomrule
\end{tabular}}
\ttabbox
{\caption{SVM and Random Forest Training Configurations}\label{table:classic_params}}{
\begin{tabular}{@{}ll@{}}
\toprule
Classifier     & Parameters                      \\ \midrule
SVM            & C: 1,10,100                     \\
               & Kernel: linear,rbf,poly,sigmoid \\
Random Forest  & Criterion: gini, entropy        \\
               & Num Estimators: 5,10,50         \\
               & Max Depth: 5,10,50              \\ \bottomrule
\end{tabular}}
\end{floatrow}
\end{table}

After training is complete on all CNN architectures and hyperparameters, the best performing CNN with respect to the validation set accuracy for both LIDAR and RGB images is selected for use in feature extraction and subsequent training with SVM and random forest classifiers.  In this scenario only the layers including and before FC1 are used to generate a condensed feature map the size of FC1 to represent the image.  The augmented training set images are \emph{reduced} to this small feature vector, and are used to train both sets of classifiers over a variety of configurations as shown in Table \ref{table:classic_params}. 
% A graphical representation is shown in Figure \ref{fig:cnn_features}.

\section{Results}

\subsection{Case Study and Data Set Generation}
The geographic region of interest chosen for analysis was the city Witten in the North Rhine-Westphalia state (NWR) of Germany.  The state of NWR has recently (May 2017) made public their full LIDAR data set with a permissible license \citep{lidardata}.  In addition the city of Witten has the desirable characteristic of having the most densely-labeled roof shapes in the OSM database, providing a robust pre-labeled data set for this paper's analysis.  Satellite images were garnered through using ArcGIS world imagery base maps with ground resolution of 0.2 meters. 

\begin{figure*}[ht!]
\centering
\includegraphics[width=.8\textwidth]{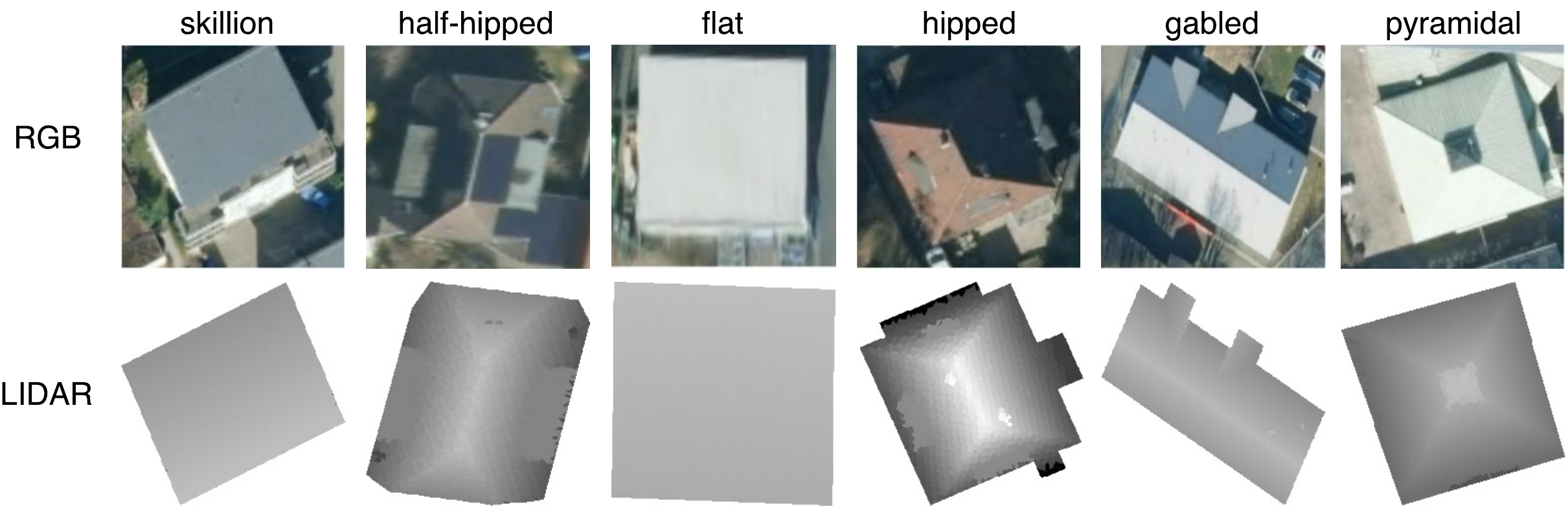}
\caption{Example of generated images}
\label{fig:all_buildings}
\end{figure*}

In the city of Witten the most prevalent roof shapes were Flat, Gabled, Hipped, Half-hipped, Skillion, and Pyramidal. All other roof shapes did not provide a sufficient number of samples (minimum 100 samples) to be used for supervised training. A total of 26,719 buildings have the above-mentioned roof shapes, and RGB and LIDAR images were generated for all buildings. Examples of generated images can be found in Figure \ref{fig:all_buildings}. To ensure an accurate data set would be used for training, validation, and testing, 2500 buildings were randomly selected for review and manually removed if any issues were detected.  Out of the 2500 buildings, 1119 were removed for reasons such as: poor LIDAR image quality, trees obscuring the roof, and the building being absent (demolished).  The remaining 1381 buildings were randomly split 60/30/10 into training, validation, and test sets, respectively.

\vspace{-0.3cm}

% A breakdown of this information can be found in Table \ref{table:buildings}. 

% \begin{table}[ht]
% \centering

% \caption{Breakdown of Witten Dataset}
% \begin{tabular}{@{}lllll@{}}
% \toprule
% Roof Shape  & All Witten & Training & Validation & Test \\ \midrule
% Flat        & 8679       & 143      & 71         & 25   \\
% Gabled      & 15572      & 185      & 92         & 32   \\
% Hipped      & 1535       & 250      & 125        & 42   \\
% Half-hipped & 505        & 124      & 62         & 22   \\
% Skillion    & 316        & 73       & 36         & 13   \\
% Pyramidal   & 112        & 51       & 25         & 10   \\ \bottomrule
% \end{tabular}
% \label{table:buildings}
% \end{table}

\subsection{CNN Training and Results}

\begin{figure}[t]

\centering
   \begin{subfigure}[b]{0.45\textwidth}
   \includegraphics[width=1\linewidth]{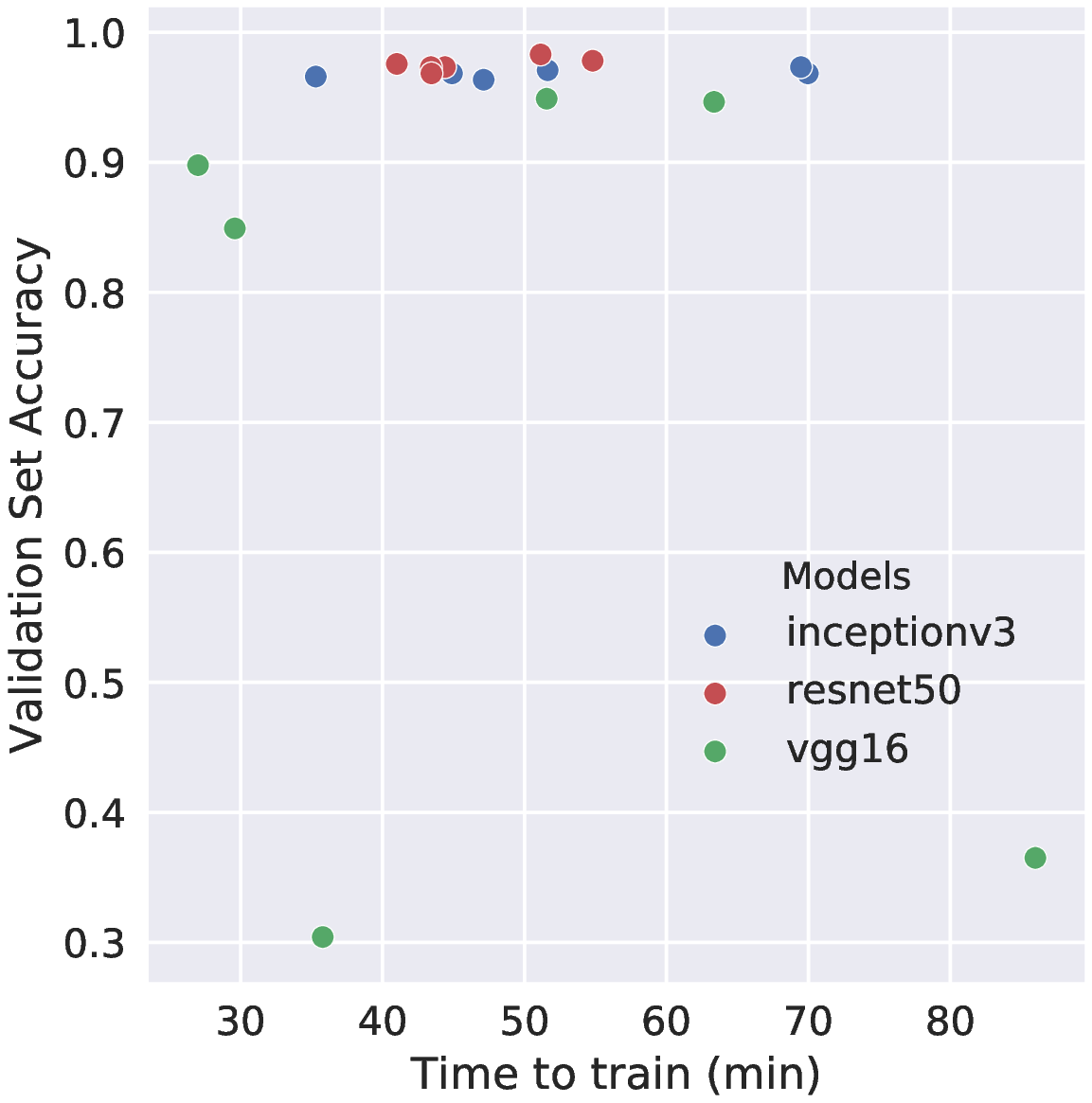}
   \caption{}
   \label{fig:cnn_lidar} 
\end{subfigure}
\begin{subfigure}[b]{.45\columnwidth}
   \includegraphics[width=1\linewidth]{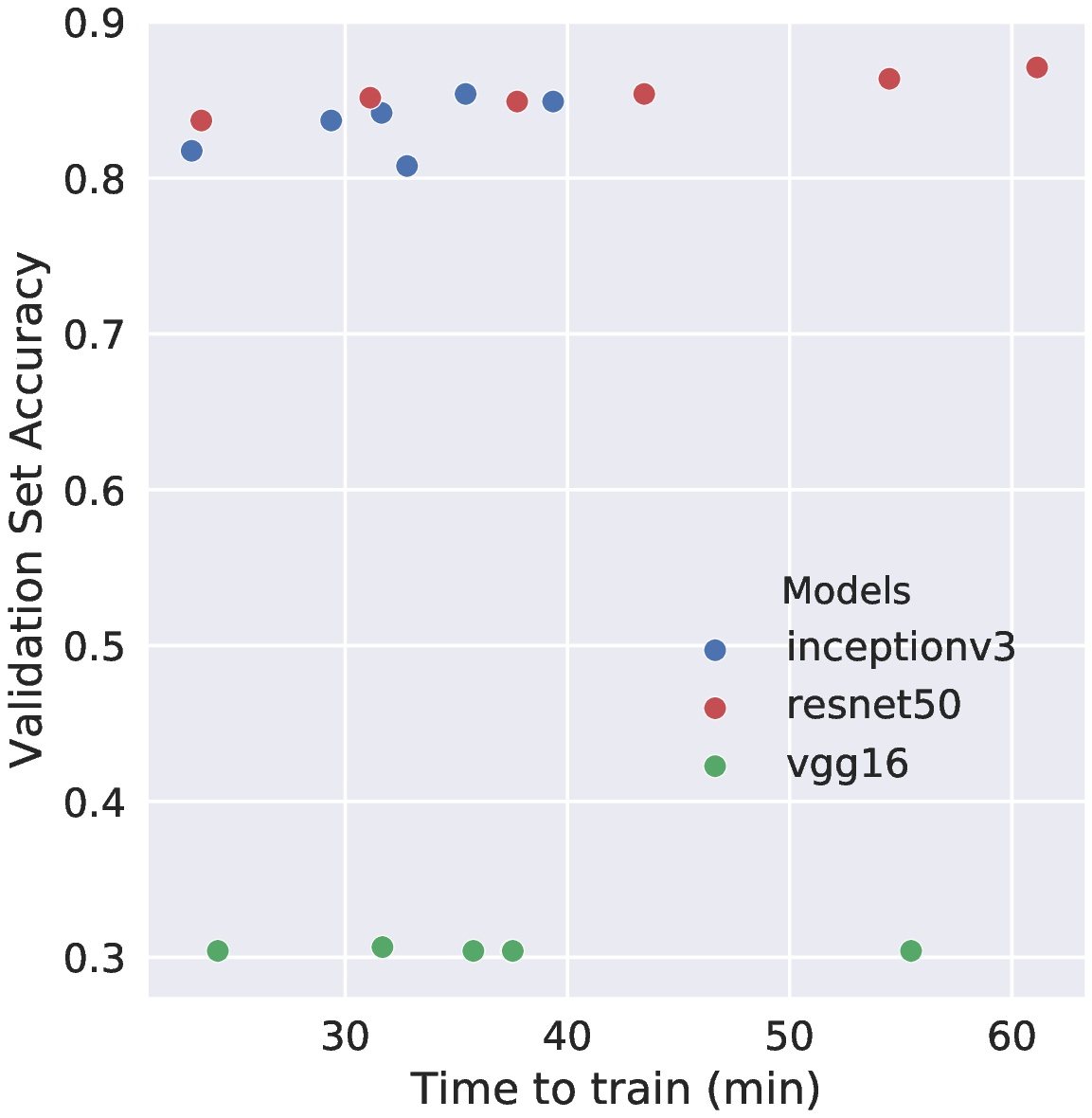}
   \caption{}
   \label{fig:cnn_rgb}
\end{subfigure}

\caption[Results of CNN training]{(a) Validation set accuracy with LIDAR input. (b) Validation set accuracy with RGB image input.}
\label{fig:cnn_all}
\end{figure}

Training was performed on the University of Michigan's Flux system, providing a server with six gigabytes of RAM, two CPU cores, and an NVIDIA Tesla K20X. A graphical overview of results over all evaluated CNN architectures and hyperparameters for LIDAR and RGB inputs can be found in Figure \ref{fig:cnn_lidar} and \ref{fig:cnn_rgb}, respectively. Each dot represents a configured training session, with the color denoting the base model architecture used, while the FC1 size and frozen layers parameters are not shown.  The vertical axis shows \emph{validation} set accuracy and the horizontal axis is the amount of time taken to train a model. 
% A full table of detailed results can be found in the Table \ref{table:cnn_all_results}

The most successful base model architectures were \texttt{Resnet50}, \texttt{Inceptionv3}, and finally \texttt{VGG16} with a relatively poor performance.  The top performing model architecture for LIDAR and RGB inputs was actually the same \texttt{Resnet50} model with a FC1 size of 100 and all 50 initial layers frozen during training. These models gave a validation accuracy of $98.3\%$ and $87.1\%$ for LIDAR and RGB inputs respectively. These results demonstrates good generalizability of the models, and leads us to select them for use in feature extraction as inputs with SVM and decision tree classifiers.

% Note that the validation set, though not directly used in training for gradient descent, is implicitly biased as it was used as an early stopping mechanism and now used to select the best performing model. 

% \begin{figure}[ht!]
% \centering
% \includegraphics[width=.8\columnwidth]{cnn_lidar_results.eps}
% \caption{Validation set accuracy with LIDAR image input} %of all CNN architectures  }
% \label{fig:cnn_lidar}
% \end{figure}

% \begin{figure}[ht!]
% \centering
% \includegraphics[width=.8\columnwidth]{cnn_rgb_results.eps}
% \caption{Validation set accuracy with RGB satellite image input}% of all CNN architectures }
% \label{fig:cnn_rgb}
% \end{figure}

\subsection{Feature Extraction and Classification Training and Testing}

\begin{figure}
\CenterFloatBoxes
\begin{floatrow}
\ffigbox
  {\includegraphics[width=1\columnwidth]{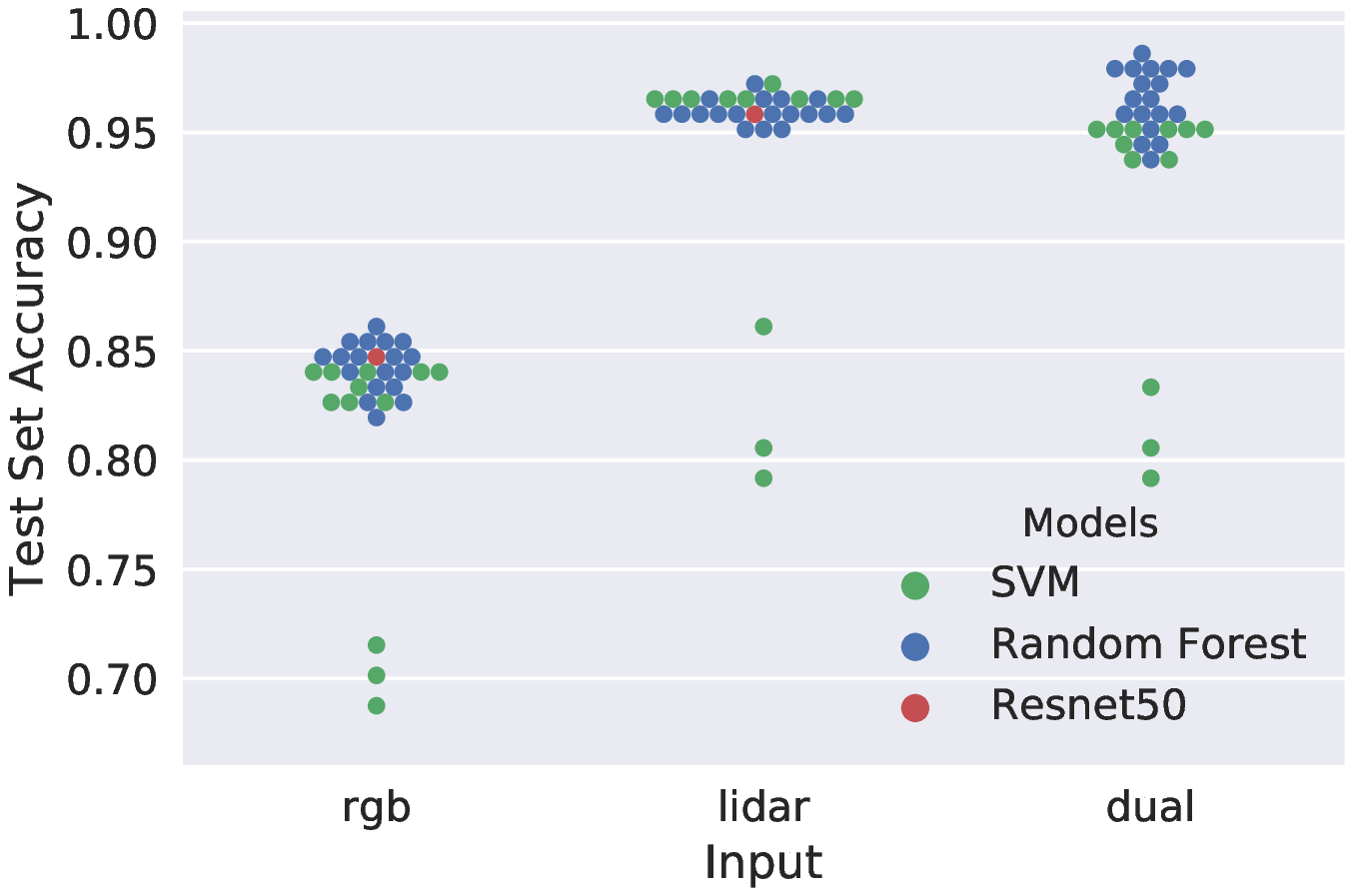}}
  {\caption{Test data set accuracies for feature extraction and training on SVM and decision tree classifiers.}
  \label{fig:feature_classical_results_test}}
\killfloatstyle
\ttabbox
  {\begin{tabular}{llll}
\hline
Input & Accuracy & Model         & Parameters                 \\ \hline
rgb   & .861          & RF & C: gini, NE: 5, MD: 50  \\
lidar & .972          & RF & C: gini, NE: 5, MD: 10  \\
dual  & .986          & RF & C: gini, NE: 50, MD: 50 \\ \hline
\end{tabular}
  }
  {\caption{Best performing feature classifiers and their associated test set accuracies. RF=Random Forest, C=Criterion, NE=Number of Estimators, MD=Maximum Depth}\label{table:features_test_result}}
\end{floatrow}
\end{figure}

The same training set is processed with the fine-tuned \texttt{Resnet50} model to produce a condensed feature set for each building image, both LIDAR and RGB.  After this processing, each building's LIDAR and RGB image is represented by a feature vector of length 100.  This new high level feature training set is then fed to SVM and random forest classifiers with varied configurations. Once all classifiers are trained, they are run against the \emph{test} data set with results shown in Figure \ref{fig:feature_classical_results_test}. The fine-tuned  \texttt{Resnet50} model is shown in red for comparison. 
% A full detail of all results can be found in the appendix.

% \begin{figure}[ht!]
% \centering
% \includegraphics[width=.8\columnwidth]{test_data_set_classical.eps}
% \caption{Results of feature extraction and training on SVM and decision tree classifiers. Accuracy of test data set shown. }
% \label{fig:feature_classical_results_test}
% \end{figure}

These results indicate that the test accuracy is able to improve modestly, around 1.4\% in the best case, by combining both the RGB and LIDAR feature sets into one fused dual input. This illustrates the ability of these algorithms to learn a nonlinear decision function that can account for the strengths and weaknesses of LIDAR and RGB data. In essence, fused data results are better than either LIDAR or RGB data alone. Table \ref{table:features_test_result} highlights the top performing models, parameters, and test set accuracies. 

\section{Discussion and Conclusions}
%An accurate database of buildings and their roof geometry has many useful applications ranging from urban mapping to emergency flight planning. 
This paper proposed fusing a combination of OpenStreetMap (OSM), satellite, and LIDAR data to generate images of buildings and predict roof shapes (types). Pre-labeled data from OSM was exploited to auto-generate a training dataset. Multiple CNN architectures were trained and tested, with the highest-performing CNN selected for image feature extraction for use with second stage SVM and decision tree classifiers. The combination of RGB and LIDAR image features support previous findings that multi-modality data fusion can improve classification accuracy. In future work, we will contribute automatically-generated roof shape labels to the OSM database. This information can be used to find safe landing sites in autonomous UAV emergency landing scenarios. Additional results and algorithm details will be presented in the full paper.

\section*{Acknowledgment}
This work was in part supported under NASA Award NNX11AO78A and NSF Award CNS 1329702.
%{\color{blue}
%In addition the authors would like to thank the many contributors of the open source software including but not limited to:

%\begin{itemize}
%     \item Python - \url{https://www.python.org/}
%     \item Rasterio - \url{https://github.com/mapbox/rasterio}
%     \item PyProj - \url{https://github.com/jswhit/pyproj}
%     \item TensorFlow - \url{https://www.tensorflow.org/}
% \end{itemize}
% }

\bibliographystyle{unsrt}
\bibliography{reference}

\end{document}